\documentclass[conference]{IEEEtran}
\usepackage{times}

\usepackage[numbers]{natbib}
\usepackage{multicol}
\usepackage[bookmarks=true]{hyperref}
\usepackage{graphicx} 
\usepackage{booktabs}
\usepackage{pifont}  
\usepackage{array}  
\usepackage{float}

\begin{document}

\title{Learning from Watching: Scalable Extraction of Manipulation Trajectories from Human Videos}




\author{\authorblockN{Xiao Hu}
\authorblockA{College of Engineering \\
Northeastern University\\
Boston, Massachusetts \\
Email: xiao.h1@northeastern.edu}
\and
\authorblockN{Gilbert Ye}
\authorblockA{College of Engineering \\
Northeastern University\\
Boston, Massachusetts \\
Email: y.ye@northeastern.edu}}


%

\maketitle

\begin{abstract}
Collecting high-quality data for training large-scale robotic models typically relies on real robot platforms, which is labor-intensive and costly, whether via teleoperation or scripted demonstrations. To scale data collection, many researchers have turned to leveraging human manipulation videos available online. However, current methods predominantly focus on hand detection or object pose estimation, failing to fully exploit the rich interaction cues embedded in these videos. In this work, we propose a novel approach that combines large foundation models for video understanding with point tracking techniques to extract dense trajectories of all task-relevant keypoints during manipulation. This enables more comprehensive utilization of Internet-scale human demonstration videos. Experimental results demonstrate that our method can accurately track keypoints throughout the entire manipulation process, paving the way for more scalable and data-efficient robot learning.
\end{abstract}

\IEEEpeerreviewmaketitle

\section{Introduction}

Scaling up robotic manipulation learning hinges on acquiring large volumes of high-quality interaction data. Conventionally, two primary strategies are employed to collect this data: teleoperation and scripted demonstrations. While teleoperation allows for expressive task execution, it demands real-time human control and the development of dedicated teleoperation interfaces for each robotic platform. These approaches are expensive, labor-intensive, and inherently difficult to scale.

The Internet offers abundant in-the-wild human manipulation videos (e.g., cooking, tool use), which are a rich yet underused resource for robotic learning. Prior works have begun leveraging such data—e.g., ViViDex \citep{chen2025vividex} tracks hands but struggles with occlusions, DexMV \citep{qin2022dexmv} relies on custom hardware, and VidBot \citep{chen2025vidbot} lacks object tracking. These approaches often miss temporal coherence or fail to capture all key entities involved.


\begin{figure}
    \centering
    \includegraphics[width=0.7\linewidth]{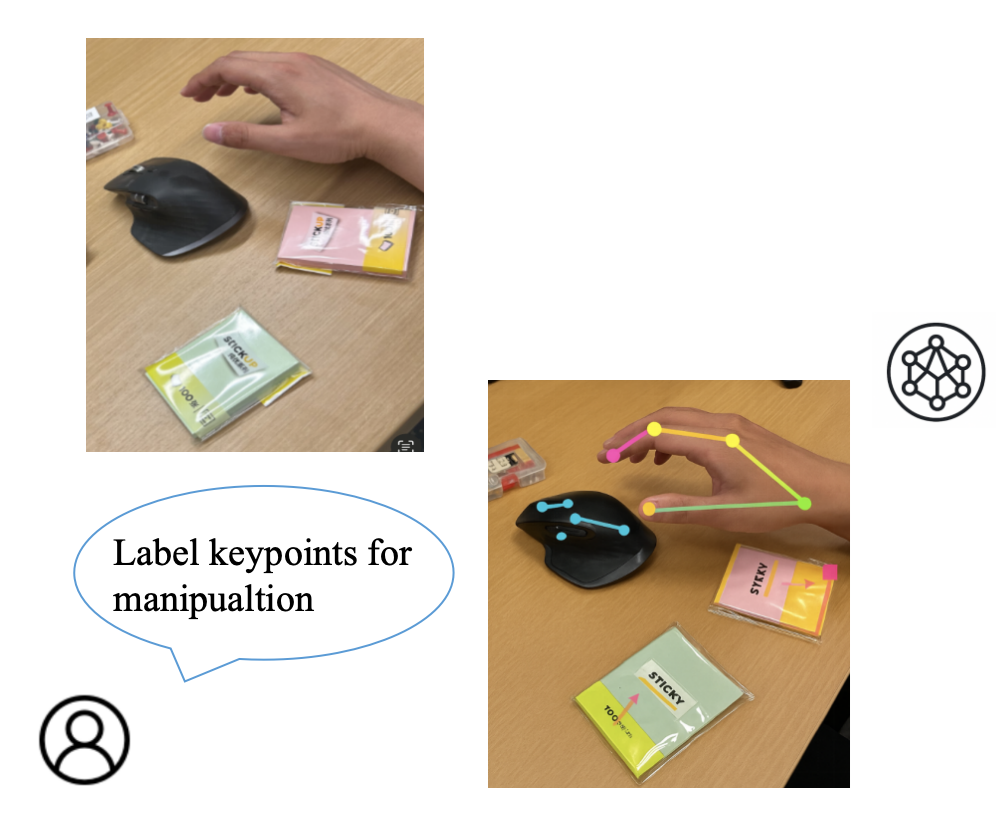}
    \label{fig:enter-label}
     \caption{\textbf{Semantic Keypoint Labeling with a Large Foundation Model}  Illustration of a large foundation model labeling keypoints for manipulation tasks. The model identifies semantic landmarks such as fingertips and tools from a real-world scene to support downstream robotic applications.} 
    \vspace{-3ex}
\end{figure}

To overcome these limitations, we propose a novel framework that combines large-scale foundation models with point tracking to extract dense and temporally consistent trajectories from raw human manipulation videos. Our approach first leverages the semantic reasoning capabilities of pretrained vision-language models to identify key stages and regions in the video. Based on this, task-relevant keypoints—such as hands, tools, and manipulated objects—are extracted and used to initialize a dense point tracking module. The core idea of our method is to bridge high-level semantic understanding with low-level pixel tracking, enabling fine-grained reconstruction of manipulation sequences from unconstrained videos.

We validate our method on a diverse collection of real-world human manipulation videos, spanning various tasks and viewpoints. Experimental results demonstrate that our framework can accurately recover complex motion trajectories of multiple keypoints with high temporal consistency and robustness. This work provides a scalable, low-cost alternative to traditional robot data collection pipelines, and offers a viable bridge between passive video demonstrations and active robotic imitation. We envision its application in scalable dataset construction, pretraining of robot policies, and advancing multimodal robot learning frameworks.

\section{Related Work}

Learning robotic manipulation from human videos is a promising way to reduce data collection costs and enhance generalization. Prior works explore this from different angles.

ViViDex~\cite{chen2025vividex} uses reinforcement learning to learn from videos without object states, but struggles under occlusion. DexMV~\cite{qin2022dexmv} translates human actions into robot policies, yet relies on custom hardware, limiting scalability. VidBot~\cite{chen2025vidbot} extracts 3D hand motions from 2D videos, but omits object dynamics essential for fine-grained tasks.

In contrast, our method leverages foundation models to guide dense keypoint tracking of hands, tools, and objects directly from unconstrained videos, enabling temporally coherent trajectory recovery without hardware or annotations.

\section{Methodology}


\begin{figure}
    \centering
    \includegraphics[width=1\linewidth]{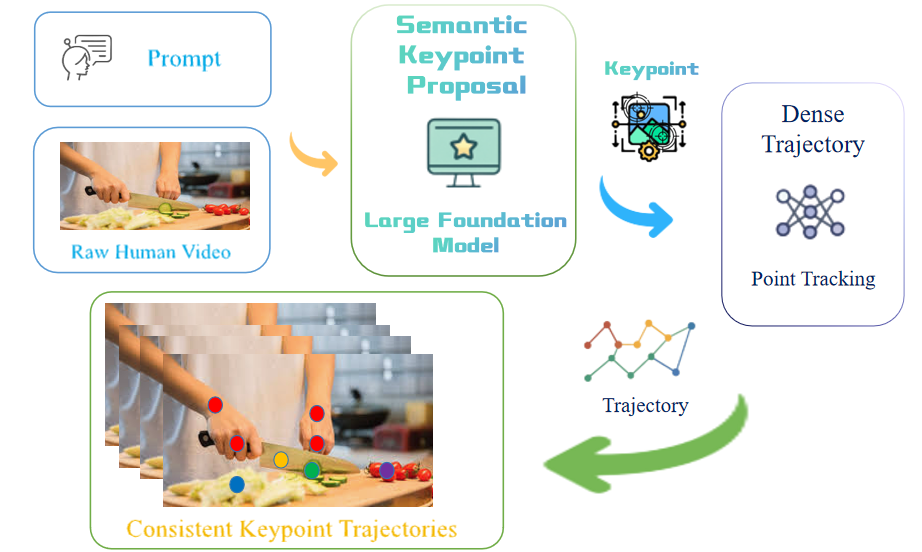}
    \caption{Overview}
    \label{fig:enter-label}
     \caption{Given a prompt and raw visual input, a large vision foundation model proposes semantic keypoints (e.g., hand joints, tools), which are then temporally tracked using a dense point tracking network to produce consistent trajectories across frames. This enables fine-grained manipulation understanding without manual annotations.} 
    \vspace{-3ex}
\end{figure}

This work aims to extract dense and temporally consistent trajectories of keypoints—including hands, tools, and manipulated objects—from in-the-wild human manipulation videos, without relying on manual annotations or additional hardware. To achieve this, we propose a two-stage trajectory extraction framework: (1) semantic keypoint proposal using a large foundation model, and (2) dense trajectory recovery via a point tracking network. Our framework integrates high-level semantic understanding with low-level pixel tracking, enabling effective reconstruction of dynamic manipulation behaviors from raw videos.

\subsection{Semantic Keypoint Proposal via Foundation Models}

Given a human manipulation video as input, we first apply a pretrained large-scale foundation model to perform semantic understanding and identify keypoints corresponding to hands, tools, and task-relevant objects. To improve detection accuracy, we select the first video frame in which clear hand presence is observed and use it as the input to guide the foundation model in locating task-relevant regions and proposing candidate keypoints.

\subsection{Trajectory generating via Point Tracking
}
Once the initial semantic keypoints are identified, we apply a dense point tracking network to recover their temporal trajectories. This module takes the video frames and the proposed keypoints as input and outputs spatially aligned trajectories across the entire sequence. This enables precise tracking of fingertips, tool tips, and object contours—even under occlusions and motion blur—resulting in a set of temporally coherent trajectories that span the full manipulation process.

To further enhance tracking stability, we employ bidirectional (forward and backward) tracking and apply consistency filtering to discard unreliable tracks. In addition, we perform trajectory interpolation for segments affected by short-term occlusions or missing detections.

\section{Experiment}

After evaluating several state-of-the-art large foundation models—GPT-4o, Gemini 1.5, Grok V, and Claude 3 Opus—we selected GPT-4o for its outstanding accuracy in recognizing keypoints annotated in publicly available datasets.

Keypoints are proposed by the large foundation model and passed to the tracking module as initial seeds. We use a pretrained LocoTrack model for point tracking, with a frame stride of 1 and a tracking window size of 64 frames. All input videos are resized to 256×256 resolution prior to inference to ensure efficient processing.

Additionally, we manually design category-specific prompts to guide the large foundation model in identifying semantically meaningful keypoints from human manipulation videos at the current stage,. To enhance the accuracy of extracted trajectories and ensure data compatibility across heterogeneous robotic platforms, we standardize the extracted motion by mapping the human wrist keypoint trajectories to the robot end-effector motion. This abstraction facilitates cross-domain transfer and enables consistent policy training regardless of the underlying embodiment.

\begin{table}[htbp]
  \centering
  \renewcommand{\arraystretch}{1.1}
  \small  
  \setlength{\tabcolsep}{3pt}  
  \begin{tabular}{lcccc}
    \toprule
    Method & Multi-obj. & Occlusion Robust. & Trajectory & Arbitrary Vid. \\
    \midrule
    ViVIDex & \ding{55} & \ding{55} & \ding{51} & \ding{51} \\
    VidBot  & \ding{55} & \ding{55} & \ding{55} & \ding{51} \\
    DexMV   & \ding{51} & \ding{51} & \ding{51} & \ding{55} \\
    Ours    & \ding{51} & \ding{51} & \ding{51} & \ding{51} \\
    \bottomrule
  \end{tabular}
  \caption{Comparison of different methods: support for multiple objects, occlusion robustness, trajectory prediction, and generalization to arbitrary videos.}
  \label{tab:method_comparison}
\end{table}



\bibliographystyle{plainnat}
\bibliography{references}

\end{document}